%% file: root.tex
\documentclass[letterpaper, 10 pt, conference]{ieeeconf}  % Comment this line out if you need a4paper

\IEEEoverridecommandlockouts                              % This command is only needed if 
\usepackage{graphics} % for pdf, bitmapped graphics files
\usepackage{amsmath} % assumes amsmath package installed
\usepackage{amssymb}  % assumes amsmath package installed
\usepackage{bm}

\usepackage{xurl}
\usepackage{booktabs}
\usepackage{multirow}
\usepackage{multicol}
\usepackage{enumerate}
\usepackage{booktabs}
\usepackage{soul}
\usepackage{xcolor}
\usepackage{cleveref}
\usepackage{graphicx}
\usepackage{caption}
\usepackage{subcaption}

\usepackage[mode=text]{siunitx}
\usepackage{physics}
\AtBeginDocument{\RenewCommandCopy\qty\SI}

\usepackage{cite}

\usepackage[acronym,style=super,nogroupskip,nonumberlist,toc=false]{glossaries-extra}
\usepackage{acro}
\input{misc/acronyms}

\crefname{equation}{Eq.}{Eq.}
\Crefname{equation}{Eq.}{Eq.}
\crefname{figure}{Fig.}{Figs.}
\Crefname{figure}{Fig.}{Figs.}

\newcommand{\mat}[1]{\mathbf{#1}}
\newcommand{\vect}[1]{\boldsymbol{#1}}
\newcommand{\quat}[1]{\mathbf{#1}}

\newcommand{\updated}[1]{{#1}}

\title{\LARGE \bf
Bridging the Basilisk Astrodynamics Framework with ROS 2 for Modular Spacecraft Simulation and Hardware Integration
}

\author{E. Krantz, N. N. Chan, G. Tibert, H. Mao, and C. Fuglesang%
\thanks{This work was partially supported by the Wallenberg AI, Autonomous Systems and Software Program (WASP) funded by the Knut and Alice Wallenberg Foundation.}%
\thanks{All authors are with the KTH Royal Institute of Technology, Stockholm, Sweden. Emails: \texttt{\{eliaskra, channn, tibert, huina, cfug\}@kth.se}}%
}

\begin{document}

\maketitle
\bstctlcite{IEEEexample:BSTcontrol}
\thispagestyle{empty}
\pagestyle{empty}

%%%%%%%%%%%%%%%%%%%%%%%%%%%%%%%%%%%%%%%%%%%%%%%%%%%%%%%%%%%%%%%%%%%%%%%%%%%%%%%%
\begin{abstract}
Integrating high-fidelity spacecraft simulators with modular robotics frameworks remains a challenge for autonomy development. This paper presents a lightweight, open-source communication bridge between the Basilisk astrodynamics simulator and the Robot Operating System~2 (ROS~2), enabling real-time, bidirectional data exchange for spacecraft control. The bridge requires no changes to Basilisk's core and integrates seamlessly with ROS~2 nodes. We demonstrate its use in a leader–follower formation flying scenario using nonlinear model predictive control, deployed identically in both simulation and on the ATMOS planar microgravity testbed. This setup supports rapid development, hardware-in-the-loop testing, and seamless transition from simulation to hardware. The bridge offers a flexible and scalable platform for modular spacecraft autonomy and reproducible research workflows.
\end{abstract}

%%%%%%%%%%%%%%%%%%%%%%%%%%%%%%%%%%%%%%%%%%%%%%%%%%%%%%%%%%%%%%%%%%%%%%%%%%%%%%%%

\input{sections/SW-Releases}
\input{sections/Introduction}

\input{sections/Related_Works}

\input{sections/Implementations}
\input{sections/Dynamics_and_Control}
\input{sections/Validation}
\input{sections/Discussion_Conclusions}

% \section*{Acknowledgment}
% This work was partially supported by the Wallenberg AI, Autonomous Systems and Software Program (WASP) funded by the Knut and Alice Wallenberg Foundation.

% \clearpage
\bibliographystyle{IEEEtran}
\bstctlcite{IEEEexample:BSTcontrol}
\bibliography{References/ieeectl,References/channn_ref,References/eliaskra_ref}

\end{document}

%% file: misc/acronyms.tex
\DeclareAcronym{LQR}{
  short = LQR,
  long  = linear quadratic regulator,
  short-indefinite = an,
  long-indefinite = a
}

\DeclareAcronym{SITL}{
  short = SITL,
  long  = software-in-the-loop,
  short-indefinite = a,
  long-indefinite = a
}

\DeclareAcronym{HITL}{
  short = HITL,
  long  = hardware-in-the-loop,
  short-indefinite = a,
  long-indefinite = a
}

\DeclareAcronym{FMU}{
  short = FMU,
  long  = flight management unit,
  short-indefinite = a,
  long-indefinite = a
}

\DeclareAcronym{CA}{
  short = CA,
  long  = control allocator,
  short-indefinite = a,
  long-indefinite = a
}

\DeclareAcronym{RTAI}{
  short = RTAI,
  long  = real-time application interface,
  short-indefinite = a,
  long-indefinite = a
}

\DeclareAcronym{DOF}{
  short = DoF,
  long  = degrees-of-freedom,
  short-indefinite = a,
  long-indefinite = a
}

\DeclareAcronym{mocap}{
  short = MoCap,
  long  = motion capture system,
  short-indefinite = a,
  long-indefinite = a
}

\DeclareAcronym{MPC}{
  short = MPC,
  long  = Model Predictive Control,
  short-indefinite = an,
  long-indefinite = a
}

\DeclareAcronym{NMPC}{
  short = NMPC,
  long  = Nonlinear Model Predictive Control,
  short-indefinite = an,
  long-indefinite = a
}

\DeclareAcronym{PWM}{
  short = PWM,
  long  = pulse width modulation,
  short-indefinite = a,
  long-indefinite = a
}

\DeclareAcronym{EKF}{
  short = EKF,
  long  = extended Kalman filter,
  short-indefinite = an,
  long-indefinite = a
}

\DeclareAcronym{RTOS}{
  short = RTOS,
  long  = real-time operating system,
  short-indefinite = an,
  long-indefinite = a
}

\DeclareAcronym{ISS}{
  short = ISS,
  long  = International Space Station,
  short-indefinite = an,
  long-indefinite = a
}

\DeclareAcronym{FSW}{
  short = FSW,
  long  = flight software,
  short-indefinite = an,
  long-indefinite = a
}

\DeclareAcronym{SFF}{
  short = SFF,
  long  = spacecraft formation flying,
  short-indefinite = an,
  long-indefinite = a
}

\DeclareAcronym{LEO}{
  short = LEO,
  long  = Low-Earth-Orbit,
  short-indefinite = a,
  long-indefinite = a
}

\DeclareAcronym{GNC}{
  short = GNC,
  long  = {Guidance, Navigation \& Control},
  short-indefinite = a,
  long-indefinite = a
}

\DeclareAcronym{ROS2}{
  short = ROS~2,
  long  = Robot Operating System 2,
  short-indefinite = a,
  long-indefinite = a
}

\DeclareAcronym{cFS}{
  short = cFS,
  long  = core Flight System,
  short-indefinite = a,
  long-indefinite = a
}

\DeclareAcronym{ZMQ}{
  short = ZMQ,
  long  = ZeroMQ,
  short-indefinite = a,
  long-indefinite = a
}

\DeclareAcronym{ECI}{
  short = ECI,
  long  = Earth-Centered Inertial,
  short-indefinite = an,
  long-indefinite = a
}

\DeclareAcronym{CW}{
  short = CW,
  long  = Clohessy-Wiltshire,
  short-indefinite = a,
  long-indefinite = a
}

\DeclareAcronym{MRPs}{
  short = MRPs,
  long  = Modified Rodrigues Parameters,
  short-indefinite = an,
  long-indefinite = a
}

\DeclareAcronym{OCP}{
  short = OCP,
  long  = Optimal Control Problem,
  short-indefinite = an,
  long-indefinite = a
}

%% file: sections/SW-Releases.tex
\section*{Software Releases}

% \begin{itemize}
%     \item \textit{BSK-ROS2-Bridge} \\
%     \url{https://github.com/DISCOWER/bsk-ros2-bridge}
    
%     \item \textit{Basilisk ROS 2 Messages} \\
%     \url{https://github.com/DISCOWER/bsk-msgs}
    
%     \item \textit{Demo MPC using BSK-ROS2-Bridge} \\
%     \url{https://github.com/DISCOWER/bsk-ros2-mpc}
% \end{itemize}

\begin{itemize}
    \item \textit{BSK-ROS2-Bridge} \\
    {\footnotesize\texttt{https://github.com/DISCOWER/bsk-ros2-bridge}}
    
    \item \textit{Basilisk ROS 2 Messages} \\
    {\footnotesize\texttt{https://github.com/DISCOWER/bsk-msgs}}
    
    \item \textit{Demo MPC using BSK-ROS2-Bridge} \\
    {\footnotesize\texttt{https://github.com/DISCOWER/bsk-ros2-mpc}}
\end{itemize}

%% file: sections/Introduction.tex
\section{Introduction}

\Ac{SFF} enables advanced mission architectures by coordinating multiple spacecraft as a distributed system. This paradigm supports modular assembly, distributed sensing, and in-orbit inspection, but also demands precise control and inter-agent coordination under tight dynamic constraints~\cite{scFormationControllerStateOfTheArt}. Demonstration missions such as PRISMA~\cite{PRISMA-mission} and PROBA-3~\cite{PROBA-3-Mission} illustrate the increasing performance requirements, with recent efforts targeting sub-meter accuracy. As autonomy becomes more critical, so does the ability to validate and iterate on control architectures quickly and reliably across different test environments.

High-fidelity simulators like Basilisk (BSK)~\cite{avs_laboratory_basilisk_nodate} are essential for modeling spacecraft dynamics, multi-body interactions, and orbital environments in autonomy development workflows. Meanwhile, \ac{ROS2}~\cite{ROS2} has become the standard for modular, real-time capable robotics software, supported by a broad ecosystem for autonomy development. \updated{Bridging Basilisk's orbital dynamics and flight-software interfaces with \ac{ROS2}'s distributed middleware enables a unified autonomy pipeline across simulation, \ac{SITL}, and \ac{HITL}, addressing the lack of native compatibility and providing orbital fidelity unavailable in general-purpose robotic simulators.}

\subsection*{Contributions}

This paper introduces an open-source bridge between Basilisk and \ac{ROS2}, facilitating bidirectional communication between high-fidelity spacecraft simulations and modular robotics software. Our main contributions are:

\begin{itemize}
    \item A lightweight and scalable communication interface connecting \ac{ROS2} nodes with Basilisk simulations without modifying Basilisk's internal architecture.
    \item Demonstration of the proposed bridge in a leader-follower \ac{SFF} scenario using \ac{NMPC}, fully implemented in \ac{ROS2}.
    \item Comparative validation in both simulation and hardware using the KTH ATMOS free-flyers~\cite{roque_towards_2025}, showing consistent control performance across platforms.
\end{itemize}

This work aims to support rapid prototyping and reliable software transition across simulation and hardware systems.

%% file: sections/Related_Works.tex
\section{Related Work}

The Basilisk Astrodynamics Simulation Framework provides a modular environment combining a Python scripting interface with a high-performance C/C++ backend, \updated{widely used for spacecraft \ac{FSW} development~\cite{basiliskIntro}}. Features like accelerated time, visualization via the Vizard engine, and, recently, the integration with MuJoCo~\cite{todorov2012mujoco} \updated{extend its applicability to articulated-bodies and space robotics~\cite{Bonilla:2025kh}.} However, integrating external autonomy software with Basilisk remains non-trivial. \updated{Typical workflows require implementing autonomy modules within Basilisk's internal messaging system, posing a steep learning curve. Prior bridging efforts, such as the ROS–Basilisk interface in~\cite{Basilisk-ROS2-Bridge-1}, embed \ac{ROS2} execution within Basilisk through timer-coupled operation and per-message translation, which complicates time synchronization and increases maintenance overhead as message definitions and \ac{ROS2} versions evolve. Similar integrations, such as with NASA's \ac{cFS}~\cite{kramlich_validation_2023}, have demonstrated feasibility but remain tightly coupled to specific software versions.} In contrast, general-purpose robotic simulators like Gazebo~\cite{koenig_design_2004} and Isaac Sim~\cite{nvidia_corporation_isaac_nodate} offer native \ac{ROS2} support but lack the orbital dynamics fidelity required for spacecraft scenarios.

\Ac{ROS2} itself has become a widely adopted framework for modular, distributed autonomy, supported by its DDS communication layer. Its design makes it well-suited for integrating complex robotic systems, including those in space environments. NASA's Astrobee free-flyers aboard the ISS exemplify its use for onboard navigation, control, and perception~\cite{fluckiger_astrobee_2018}. Recent initiatives such as Space ROS~\cite{spaceROS}, a safety-focused fork of \ac{ROS2} developed by NASA and collaborators, further underscore the framework's relevance for mission-critical space robotics.

To enable representative testing of autonomous spacecraft systems, ground-based testbeds have become essential. These platforms provide repeatable microgravity-analog conditions, enabling \ac{HITL} testing of planning and control algorithms. Notable examples include NASA Ames' granite-based facility for Astrobee~\cite{bualat_astrobee_2015}, Caltech's large-scale resin-floor testbed~\cite{barrett_demonstration_2009}, and ESA's Orbital Robotics Laboratory~\cite{kolvenbach_recent_2016}. At KTH Royal Institute of Technology, the ATMOS platform~\cite{roque_towards_2025} offers an open-source 2D-spacecraft analog with air-based propulsion. It uses PX4 Autopilot~\cite{meier_px4_2015} for low-level control and \ac{ROS2} for high-level autonomy, supporting \ac{HITL} evaluation of spacecraft control and planning algorithms under realistic and repeatable conditions.

%% file: sections/Implementations.tex
\section{Basilisk-ROS~2 Bridge}

The \textit{BSK-ROS2-Bridge} enables modular spacecraft autonomy development by connecting Basilisk's high-fidelity simulation environment with the \ac{ROS2} middleware. This open-source interface allows seamless and scalable data exchange between Basilisk and \ac{ROS2}, supporting both real-time and accelerated simulation workflows. The bridge is implemented as a \ac{ROS2} package and leverages Basilisk's existing support for \ac{ZMQ}~\cite{imatix_zeromq_nodate}, a lightweight messaging protocol used internally by Basilisk's visualization tool Vizard. By integrating with Basilisk's native message infrastructure without modifying its core, the bridge enables autonomy stacks to run externally in \ac{ROS2} while maintaining synchronized interaction with the simulation. This design lowers the barrier to entry for researchers and engineers accustomed to robotics middleware, and supports multi-agent, real-time, and \ac{HITL} scenarios with minimal setup.

\subsection{Architecture}

\begin{figure}[tb]
    \centering
    \includegraphics[width=0.85\linewidth]{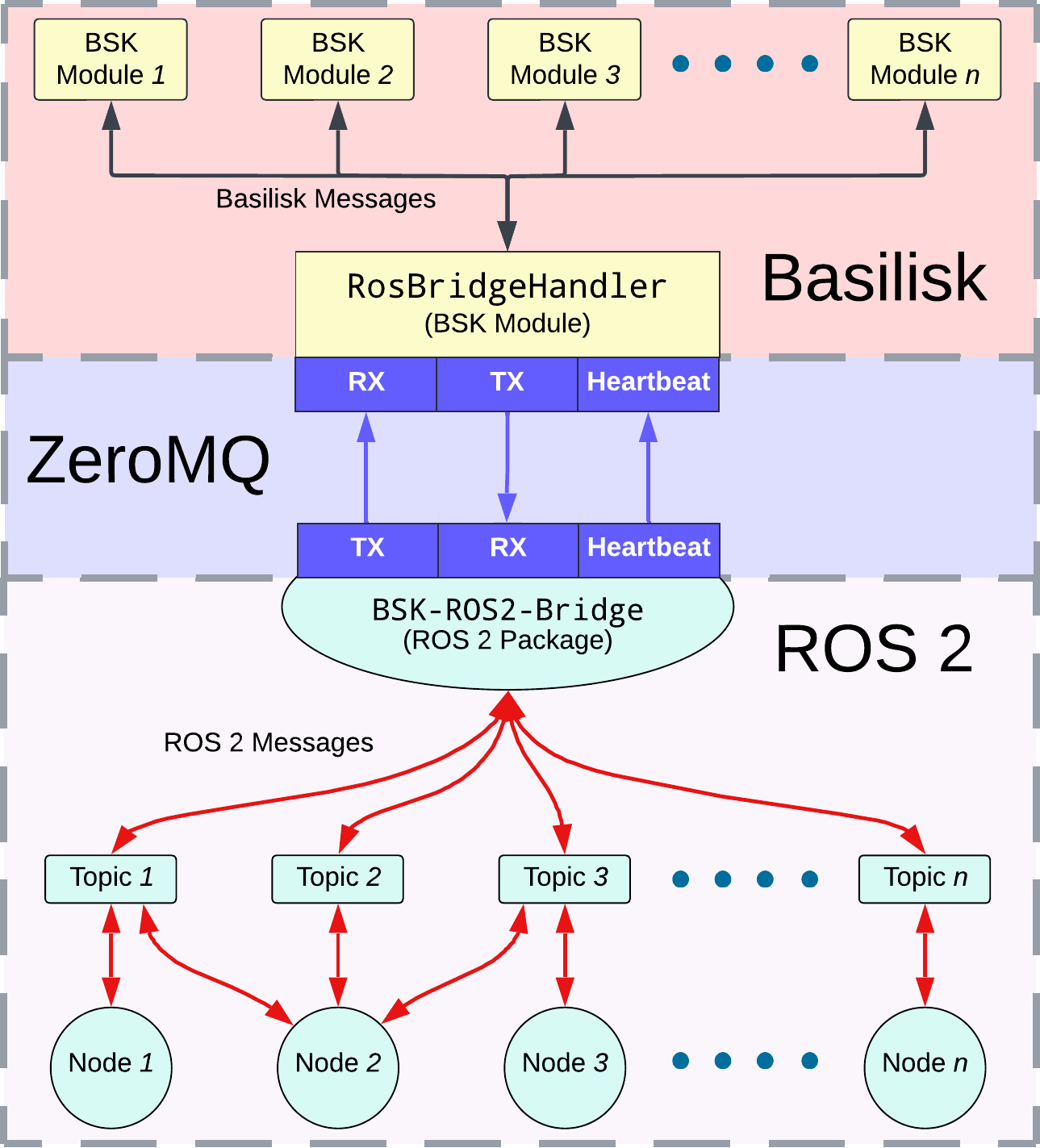}
    \caption{\updated{Functional overview of the Basilisk – \ac{ROS2} bridge. The bridge interfaces with the \ac{ROS2} network and exchanges data with a Basilisk handler module, which connects messages to internal simulation modules.}}
    \label{fig:bsk-ros2-bridge-flowchart}
\end{figure}

The bridge architecture consists of two main components: a \ac{ROS2} node and a Basilisk-side handler module, \cref{fig:bsk-ros2-bridge-flowchart}. The bridge node behaves like any other \ac{ROS2} component, integrating naturally into launch files and workflows. It operates independently and can remain active across multiple simulations, communicating externally with Basilisk via three \ac{ZMQ} ports: receive, transmit, and heartbeat \updated{(used for monitoring connection health).} On the Basilisk side, the handler module follows the standard Python module structure and is imported into the simulation script like any other Basilisk module. This design allows users to extend existing scenarios with \ac{ROS2} connectivity by simply adding the handler to their simulation configuration. \updated{The integration is designed for accessibility: message routing is defined in the Basilisk script, while the bridge node requires no manual topic configuration in \ac{ROS2}, keeping a minimal setup.}

Communication between the bridge node and the Basilisk handler occurs via JSON-formatted byte messages over \ac{ZMQ} sockets, supporting low-latency and high-frequency control loops. Additionally, the bridge publishes the Basilisk simulation time to \ac{ROS2} via the \texttt{/clock} topic, allowing all \ac{ROS2} nodes to remain synchronized with the simulation time, enabling consistent behavior across nodes and supporting both real-time and accelerated simulation modes.

\subsection{Scalability and Message Configuration}

The bridge supports both single- and multi-spacecraft scenarios. A single bridge and handler instance can manage any number of spacecraft using namespace-aware topic conventions. Message routing is configured programmatically in the Basilisk simulation script, where users register publishers and subscribers via the handler's Python API by specifying the \ac{ROS2} topic name, message type, internal identifier, and namespace. Topics follow the structure \texttt{/[\textit{ns}]/bsk/[\textit{direction}]/[\textit{topic}]}, where \texttt{[\textit{direction}]} is either \texttt{in} (for \ac{ROS2}~$\rightarrow$~Basilisk) or \texttt{out} (for Basilisk~$\rightarrow$~\ac{ROS2}). This convention enables scalable, multi-agent setups and integration with \ac{HITL} experiments.

To further simplify integration, Basilisk's native message types are systematically converted into \ac{ROS2}-compatible message definitions, maintained in a standalone Git repository (\textit{bsk-msgs}). This repository includes tools to automatically update messages, ensuring continuous compatibility with future Basilisk updates. High-throughput applications are supported by launching multiple bridge-handler pairs on separate ports, yielding parallel communication streams suitable for large-scale formations, \ac{HITL} testing, and high-rate simulations.

%% file: sections/Dynamics_and_Control.tex
\section{Formation Control Demonstration}\label{sec:demo}

\updated{To benchmark the BSK-ROS2 bridge in a coordinated multi-agent setting, we implement a formation flying scenario with one \textit{leader} and two \textit{follower} spacecraft.} Each is controlled via a \ac{NMPC} in \ac{ROS2}, with the followers maintaining a desired relative formation as the leader moves between waypoints.

\subsection{Dynamics Modeling in Simulation}

\begin{figure}[tb]
    \centering
    \includegraphics[width=0.91\linewidth]{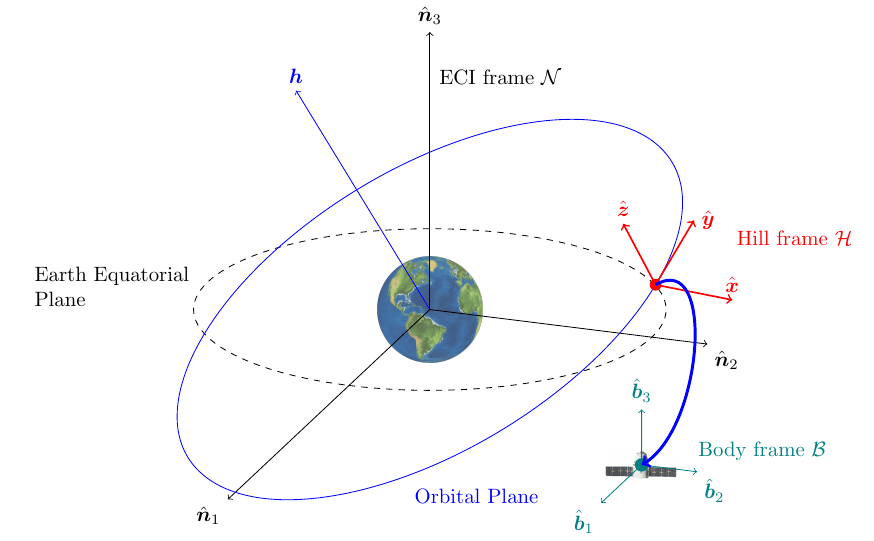}
    \caption{\updated{Hill frame $\mathcal{H}$ (red), ECI frame $\mathcal{N}$ (black) and body frame $\mathcal{B}$ (green) with respect to a leader's orbit (blue)~\cite{thomas-SFF-Thesis}.}}
    \label{fig:hill-frame}
\end{figure}

\updated{The local orbital Hill frame $\mathcal{H}$ provides a convenient reference for describing relative dynamics of spacecraft in orbit (see \cref{fig:hill-frame}). This rotating reference frame is centered at a designated leader spacecraft in a nearly circular orbit.} Defined with respect to the \ac{ECI} frame $\mathcal{N}$, its axes are:
\begin{equation}
\label{eq:Hill-frame}
\hat{\vect{x}} = \frac{\vect{p}}{\| \vect{p} \|}, \quad
\hat{\vect{z}} = \frac{\vect{h}}{\| \vect{h} \|}, \quad
\hat{\vect{y}} = \hat{\vect{z}} \times \hat{\vect{x}},
\end{equation}
where $\vect{p}$ and $\vect{v}$ are the leader's position and velocity in the inertial frame, and $\vect{h} = \vect{p} \times \vect{v}$ is its specific angular momentum. This frame serves as the basis for describing the relative dynamics of a follower spacecraft with respect to the leader. 

% \updated{The follower's motion relative to the leader can be approximated using the \ac{CW} equations~\cite{CW-Eqn}. These linearized equations describe the acceleration in $\mathcal{H}$, assuming small relative distances and velocities, $\|p^\mathcal{H}\| \ll a$, where $a$ is the semi-major axis of the leader's orbit.} Expressed in $\mathcal{H}$, centered on the leader, the follower's relative position evolves according to:
% \begin{align}
% \label{eq:CW-eqn}
% \begin{split}
%     \ddot{p}_x - 2n \dot{p}_y - 3n^2 p_x &= F_x / m \ , \\
%     \ddot{p}_y + 2n \dot{p}_x &= F_y /m \ , \\
%     \ddot{p}_z + n^2 p_z &= F_z / m \ ,
% \end{split}
% \end{align}
% \updated{where $n = \sqrt{\mu/a^3}$ is the mean motion of the leader's orbit, $\mu$ is the gravitational parameter, $F_x, F_y \text{ and } F_z$ are the external force components acting on the follower, and $m$ is the follower spacecraft's mass.}

\updated{The follower's motion relative to the leader can be approximated using the \ac{CW} equations~\cite{CW-Eqn}. Expressed in $\mathcal{H}$, centered on the leader, the follower's relative position evolves according to:
\begin{align}
\label{eq:CW-eqn}
\begin{split}
    \ddot{p}_x - 2n \dot{p}_y - 3n^2 p_x &= F_x / m \ , \\
    \ddot{p}_y + 2n \dot{p}_x &= F_y / m \ , \\
    \ddot{p}_z + n^2 p_z &= F_z / m \ ,
\end{split}
\end{align}
where $n = \sqrt{\mu/a^3}$ is the mean motion of the leader's orbit, $\mu$ is the gravitational parameter, $a$ is the semi-major axis, $F_x, F_y, F_z$ are the external force components acting on the follower, and $m$ is the follower spacecraft's mass. These linearized equations are valid for small relative distances and velocities, $\|\vect{p}^\mathcal{H}\| \ll a$ and $\|\vect{v}^\mathcal{H}\| \ll na$.}

Despite their simplicity, the \ac{CW} equations reveal inherent drift and coupling effects due to Coriolis and centrifugal forces, making formation maintenance non-trivial. These fictitious forces make it challenging to maintain a stable relative configuration over time. Control strategies such as \ac{MPC} must therefore actively compensate for these effects to ensure stability, disturbance rejection, and safe separation within tight relative constraints.

The spacecraft attitude is represented using unit quaternions $\quat{q}^{\mathcal{H}|\mathcal{B}}$ and angular velocity $\vect{\omega}^\mathcal{B}$. The rotational motion is governed by:
\begin{align}
\label{eq:Rot-Dynamics}
\begin{split}
    \dot{\quat{q}}^{\mathcal{H}|\mathcal{B}} &= \frac{1}{2} \, \quat{q}^{\mathcal{H}|\mathcal{B}} \otimes \quat{q}_{\vect{\omega}^\mathcal{B}} \ ,\\
    \dot{\vect{\omega}}^\mathcal{B} &= \mat{J}^{-1} \left( \vect{\tau^\mathcal{B}} - \vect{\omega}^\mathcal{B} \times \mat{J} \vect{\omega}^\mathcal{B} \right) \ ,
\end{split}
\end{align}
where \updated{$\quat{q}_{\vect{\omega}^\mathcal{B}}$ is the quaternion representation of $\vect{\omega}_\mathcal{B}$ with zero scalar part, $\mat{J}$ is the moment of inertia matrix, $\vect{\tau^\mathcal{B}}$ is the external torque}, and $\otimes$ denotes quaternion multiplication.

\subsection{Spacecraft Formation Flying NMPC}\label{sec:mpc}

\updated{Each agent employs an \ac{NMPC} strategy where it solves an \ac{OCP} over a finite prediction horizon of \( N \) steps.} \ac{MPC} is well suited for spacecraft formation control due to its ability to handle multivariable dynamics, predict future behaviors, and enforce constraints on control effort, collision avoidance, and relative motion~\cite{rawlings_model_2020}\updated{, capabilities that standard linear controllers such as linear quadratic regulator lack. The nonlinear formulation is used as large rotational maneuvers introduce significant nonlinearities that linear controllers cannot accurately capture. At each discrete time \( k \), the controller optimizes a sequence of control inputs \( \{ \vect{u}(n|k) \}_{n=0}^{N-1} \) to minimize a cost function subject to dynamic and operational constraints, yielding a corresponding sequence of predicted states \( \{ \vect{x}(n|k) \}_{n=0}^{N} \), starting from the initial condition \( \vect{x}(0|k) = \vect{x}_k \). Only the first control input \( \vect{u}(0|k) \) is applied, and the process is repeated. For agent $a$ at time \( k \) we have the following \ac{OCP}:}
\begin{subequations}
\label{eq:nmpc_ocp}
\begin{align}
J^*(x_k) &= \min_{\vect{u}_k} \sum_{n=0}^{N-1} \ell(\vect{x}(n|k), \vect{u}(n|k)) + \ell_f(\vect{x}(N|k)) \\
\text{s.t.} \quad & \vect{x}(0|k) = \vect{x}_k \\
& \vect{x}(n+1|k) = f\big(\vect{x}(n|k), \vect{u}(n|k)\big) \\
& \vect{x}(n|k) \in \mathbb{X}, \quad \vect{u}(n|k) \in \mathbb{U} \\
& \left\| \vect{p}^{(a)}(n|k) - \vect{p}^{(b)}(n|k) \right\|_2 \geq d_{\min}, \quad \forall b \neq a
\end{align}
\end{subequations}
Here, the function \( f(\cdot) \) denotes the discrete-time dynamics model in a fixed timestep \( \Delta t \), where the state vector $\vect{x}$ and control input $\vect{u}$ are defined as \( \vect{x} = \begin{bmatrix} \vect{p}^\mathcal{H} & \vect{v}^\mathcal{H} & \mat{q}^{\mathcal{H}|\mathcal{B}} & \vect{\omega}^\mathcal{B} \end{bmatrix} \in \mathbb{R}^{13} \), and \( \vect{u} = \begin{bmatrix} \vect{F}^\mathcal{B} & \vect{\tau}^\mathcal{B} \end{bmatrix} \in \mathbb{R}^6 \).  The sets \( \mathbb{X} \subset \mathbb{R}^{13} \) and \( \mathbb{U} \subset \mathbb{R}^6 \) define admissible states and control inputs, respectively, including velocity bounds, actuation limits, and spatial constraints. The scalar \( d_{\min} \geq 0 \) defines the minimum allowable inter-agent separation distance for collision avoidance. The stage cost is defined as:
\begin{equation}
\begin{aligned}
\ell(\vect{x}, \vect{u}) = \| \vect{p} - \vect{p}_{\text{ref}} \|^2_{\mat{Q}_p}
+ \| \vect{v} - \vect{v}_{\text{ref}} \|^2_{\mat{Q}_v} \\
+ \left\| 1 - (\quat{q}^\top \quat{q}_{\text{ref}})^2 \right\|^2_{\mat{Q}_q}
+ \| \vect{\omega} - \vect{\omega}_{\text{ref}} \|^2_{\mat{Q}_\omega}
&+ \| \vect{u} \|^2_\mat{R},
\end{aligned}
\end{equation}
where \( \mat{Q} \), and \( \mat{R} \) are positive semi-definite weighting matrices. The terminal cost is defined analogously as:

\begin{equation}
\ell_f(\vect{x}(N|k)) = \left\| \vect{x}(N|k) - \vect{x}_{\text{ref}}(N|k) \right\|^2_\mat{P},
\end{equation}
with weighting matrix \( \mat{P} \). The quaternion term minimizes the scalar deviation from the desired orientation and remains well-conditioned even for large attitude changes.

In this decentralized architecture, each agent independently solves its own \ac{NMPC} problem. The leader agent tracks a reference trajectory, which may be updated at any time through new waypoints. Each follower constructs its own reference trajectory based on the leader's predicted motion, which is communicated at each control step as the solution to the leader's \ac{OCP}. Specifically, the follower applies a desired relative position and attitude to the leader's predicted state while copying the leader's velocity and angular velocity directly:
\begin{equation}
\begin{aligned}
\vect{p}_{\text{ref}}^{(f)}(n|k) &= \vect{p}^{(l)}(n|k) + \Delta \vect{p}, \\
\vect{v}_{\text{ref}}^{(f)}(n|k) &= \vect{v}^{(l)}(n|k), \\
\quat{q}_{\text{ref}}^{(f)}(n|k) &= \Delta \quat{q} \otimes \quat{q}^{(l)}(n|k), \\
\vect{\omega}_{\text{ref}}^{(f)}(n|k) &= \vect{\omega}^{(l)}(n|k),
\end{aligned}
\end{equation}
where \( \Delta \vect{p} \in \mathbb{R}^3 \) and \( \Delta \quat{q} \in \mathbb{R}^4 \) represent the desired relative formation configuration in position and attitude.

To ensure decentralized coordination, all agents communicate their predicted trajectories, i.e., their \ac{OCP} solutions, at each control step. These shared trajectories are used by other agents to enforce the pairwise collision avoidance constraints during optimization. In this way, each agent incorporates the shared future trajectories of others, enabling predictive conflict resolution and safe formation-keeping without requiring centralized planning.

%% file: sections/Validation.tex
\section{Cross-Platform Validation}

\begin{figure}[b]
    % \raggedright
    \centering
    \includegraphics[width=0.98\linewidth, trim={330 300 230 800}, clip]{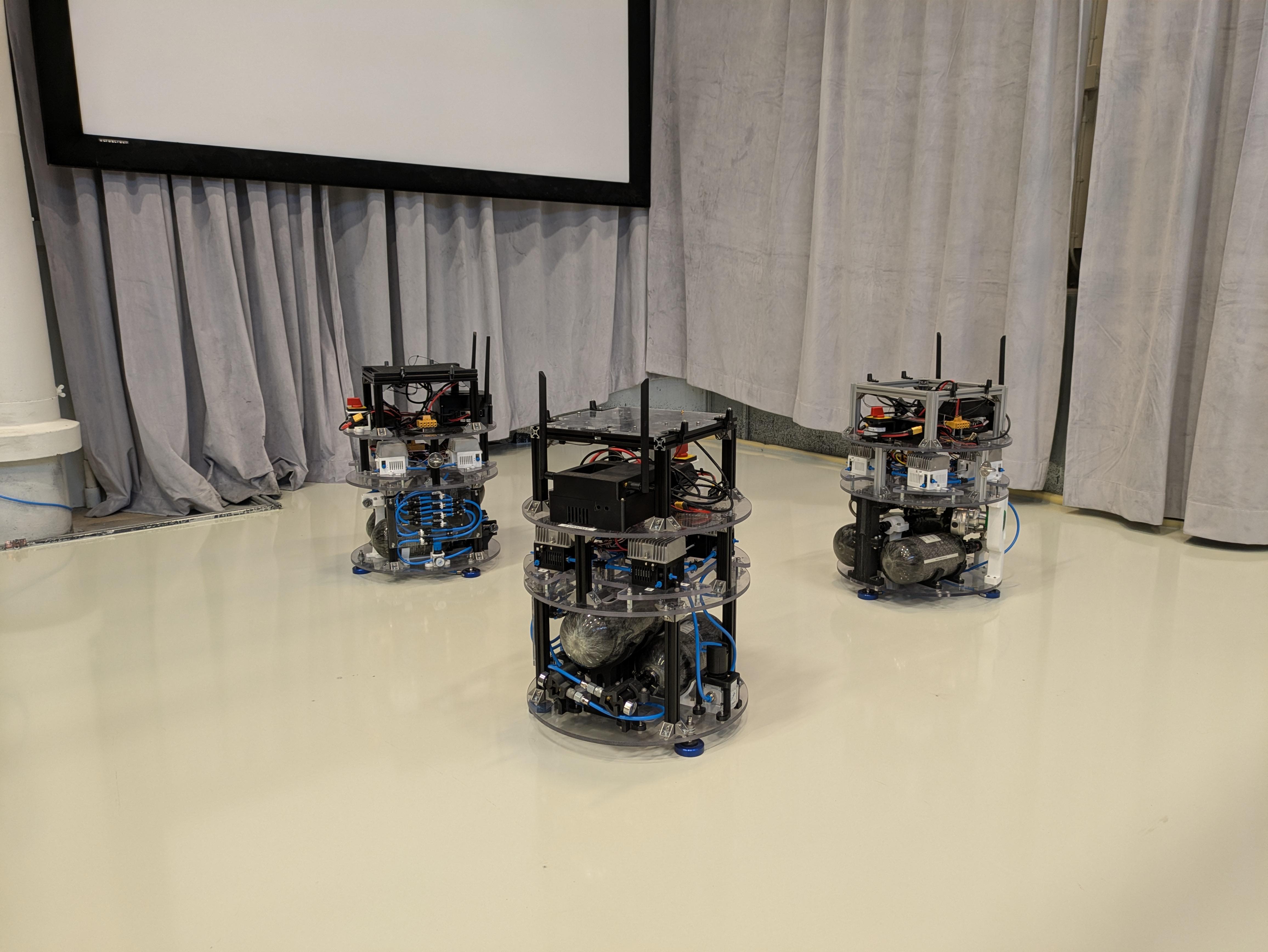}
    \caption{Three ATMOS free-flyer platforms. Each unit uses air bearings for frictionless 2D motion and is independently controlled for the spacecraft formation experiments.}
    \label{fig:Three_Atmos_photo}
    % \vspace{1.5mm}
\end{figure}

\begin{figure}[tb]
    % \raggedright
    \centering
    \includegraphics[width=0.98\linewidth, trim=280 170 170 170, clip]{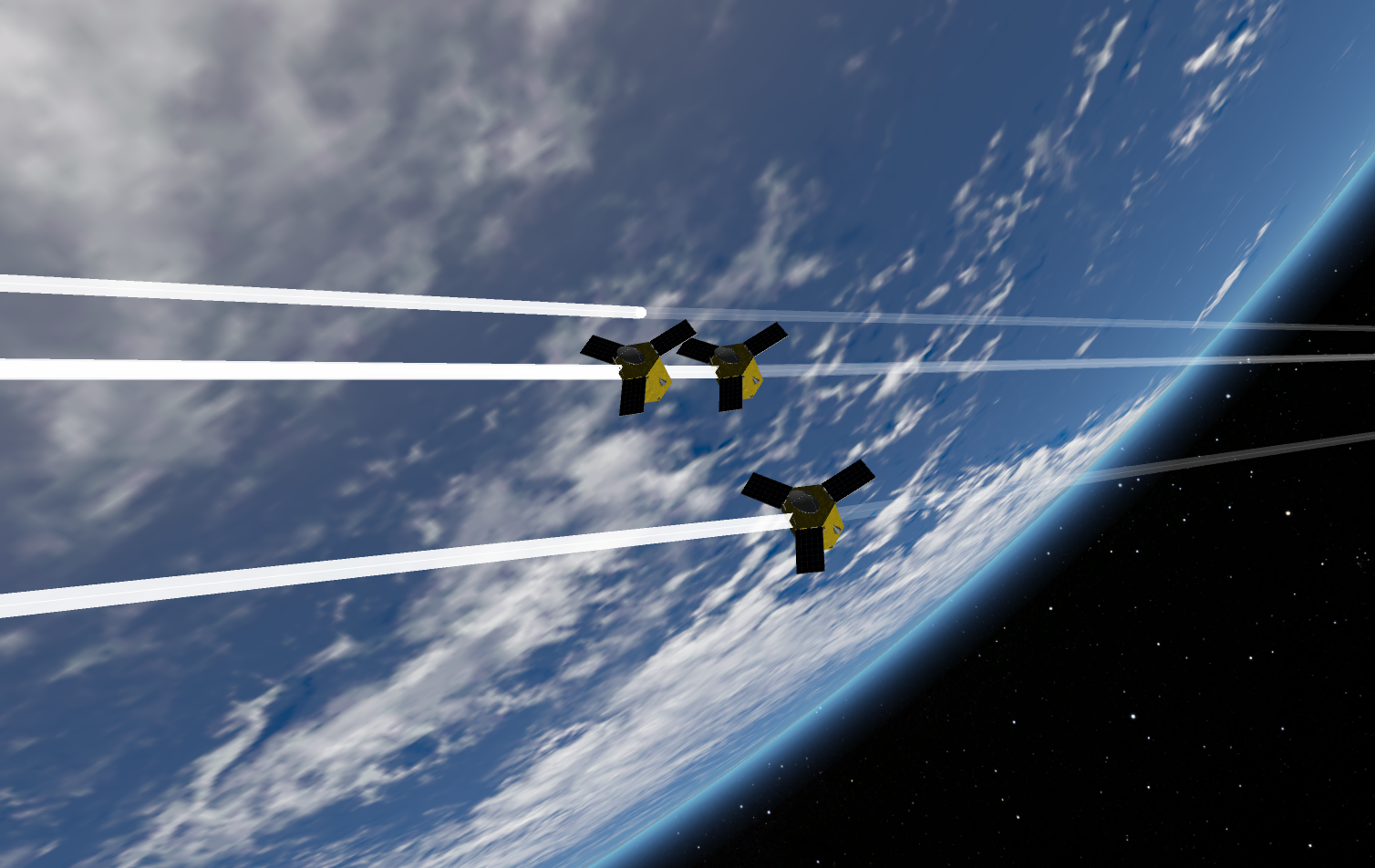}
    \caption{Basilisk simulation of the leader-follower scenario, visualized in Vizard.}
    \label{fig:vizard_formation}
\end{figure}

Using this \ac{NMPC}, we evaluate the BSK-ROS2 bridge's ability to support modular autonomy stacks by deploying an identical leader–follower \ac{SFF} in two settings: a physical hardware experiment using the planar ATMOS spacecraft analogs, \cref{fig:Three_Atmos_photo}, and a Basilisk simulation that replicates the same conditions, \cref{fig:vizard_formation}.

\subsection{Simulation and Hardware Setup}

% \updated{Both setups use the same control architecture, spacecraft parameters, and autonomy logic to ensure a consistent evaluation across environments. In each case, the full control stack operates externally as \ac{ROS2} nodes. The control strategy builds on the \ac{SFF} \ac{NMPC} controller on described in Section~\ref{sec:demo}, with three \ac{ROS2} nodes: one for the leader, which to follows a predefined waypoint sequence, and one for each follower, which maintains a fixed relative position to the leader. Each controller receives state estimates and outputs force and torque commands at \qty{5}{\hertz}, which are then allocated to actuators by the simulation or hardware stack.}

Both setups use the same control architecture, spacecraft parameters, and autonomy logic to ensure a consistent evaluation across environments. In each case, the full control stack operates externally as \ac{ROS2} nodes. The control strategy builds on the \ac{SFF} \ac{NMPC} controller described in Section~\ref{sec:demo}, with three \ac{ROS2} nodes deployed: one for the leader, which follows a predefined waypoint sequence, and one for each follower, which maintains a fixed relative position with respect to the leader. Each controller receives state estimates and outputs force and torque commands at \qty{5}{\hertz}, which are then allocated to actuators by the corresponding simulation or hardware stack.

The ATMOS free-flyer platforms provide a microgravity-analog environment for two-dimensional motion, utilizing air bearings for nearly frictionless motion over an epoxy-coated flat floor. Each unit is equipped with eight solenoid thrusters, powered by compressed air and arranged for holonomic 3-\ac{DOF} motion control (planar $\hat{\vect{x}}$, $\hat{\vect{y}}$, and yaw $\vect{\theta}_z$). The Basilisk simulation mimics this hardware configuration with three spacecraft in \ac{LEO}, parameterized to match the ATMOS platforms in mass, inertia, and actuation setup. Due to this shared hardware constraint, only thrusters are used for actuation, excluding alternatives such as reaction wheels. \updated{In simulation, the thruster configuration is expanded to twelve symmetrically arranged units, enabling full 6-\ac{DOF} actuation in 3D space while maintaining the same thruster-based actuation principle and hardware-matched limits.} Both the physical and simulated spacecraft have a mass of \qty{17.8}{\kilogram} and a moment of inertia of $J_{zz} = \qty{0.315}{\kilogram\metre\squared}$, with the simulation assuming symmetry $J_{xx} = J_{yy} = J_{zz}$ for simplicity. The maximum available thrust in $\vect{x}^\mathcal{B}$ and $\vect{y}^\mathcal{B}$ is \qty{3.0}{\newton}, and the maximum torque about $\vect{\theta}^\mathcal{B}_z$ is \qty{0.51}{\newton\metre}. 

\begin{figure*}[tpb]
    % \centering
    \begin{subfigure}[t]{0.44\textwidth}
        \raggedright
        \includegraphics[width=\linewidth]{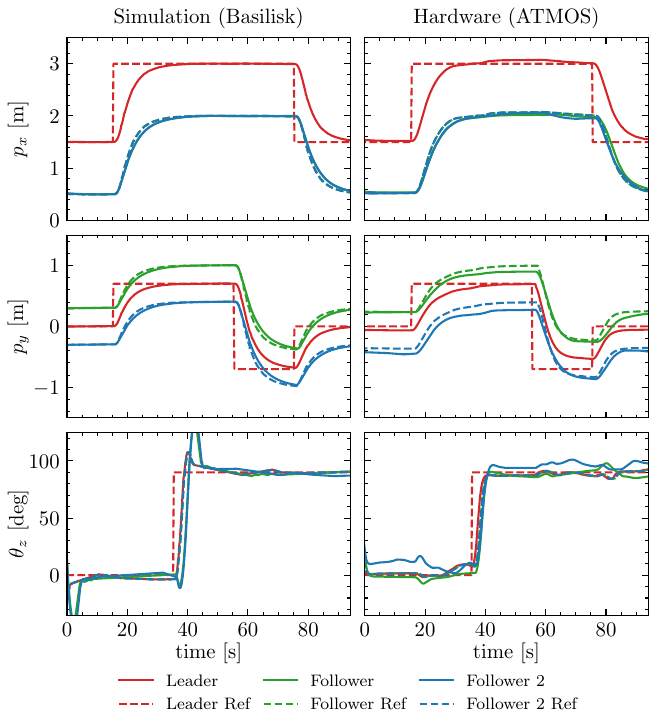}
        \caption{Position and attitude tracking in Basilisk simulation (left) and on ATMOS free-flyers (right).}
        \label{fig:bsk_px4_combined_pose_plot}
    \end{subfigure}
    \hspace{0.06\textwidth}
    \begin{subfigure}[t]{0.44\textwidth}
        \raggedright
        \includegraphics[width=\linewidth, trim=0 -15 0 0, clip]{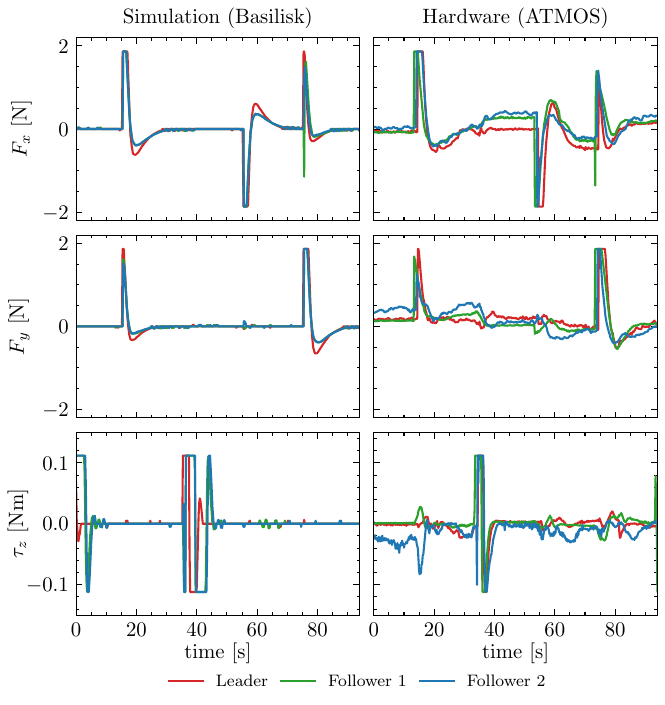}
        \caption{Commanded force and torque in Basilisk simulation (left) and on ATMOS free-flyers (right).}
        \label{fig:bsk_px4_combined_wrench_plot}
    \end{subfigure}
    \caption{Comparison of position and attitude tracking, as well as control behavior, between Basilisk simulation and ATMOS hardware in the leader–follower formation flying scenario.}
    \label{fig:pose_and_command_comparison}
\end{figure*}

In both the hardware and simulation scenarios, thrusters are actuated via \ac{PWM} at \qty{10}{\hertz}, with each thruster producing approximately \qty{1.5}{\newton} of thrust when active. For ATMOS, the controller runs offboard and transmits desired force and torque commands to the onboard PX4 flight controller, which performs control allocation via its internal mixer. In the simulation, control commands are mapped to individual thruster inputs using Basilisk's \texttt{forceTorqueThrForceMapping} module, subject to a minimum on-time of \qty{1}{\milli\second} to mimic hardware limitations.

The leader spacecraft follows a cyclic pose trajectory composed of predefined waypoints, transitioning every \qty{20}{\second}. Followers are tasked with maintaining fixed positions of \qty{-1.0}{\metre} in $\vect{x}^\mathcal{H}$ and \qty{\pm0.3}{\metre} in $\vect{y}^\mathcal{H}$ relative to the leader, while aligning their attitude with that of the leader. In the simulation case, the spacecraft are placed in a near-circular \ac{LEO}, representative of typical formation-flying mission conditions, with semi-major axis $a=\ $\qty{6778.0}{\kilo\metre}, eccentricity $e = 0.001$, inclination $i = 45^\circ$, right ascension of the ascending node $\Omega = 270^\circ$, and argument of periapsis $\omega = 90^\circ$. This choice provides realistic dynamic conditions while remaining general enough for other orbital regimes, and it does not affect comparability with the ATMOS hardware results, which are constrained to planar motion. 

Both scenarios share an identical \ac{NMPC} formulation, solved in real-time using the ACADOS solver~\cite{verschueren_acadosmodular_2022}. The prediction horizon is set to $N = 30$ with timestep $\Delta t = \qty{0.2}{\second}$. The cost function penalizes position, velocity, attitude, and angular velocity errors using a diagonal state weighting matrix $\mat{Q} = \mathrm{diag}(1 \mat{I}_3,\ 30 \mat{I}_3,\ 1000,\ 10 \mat{I}_3)$, control effort via $\mat{R} = \mathrm{diag}(0.2 \mat{I}_3,\ 100 \mat{I}_3)$, and a terminal cost via $\mat{P} = 20 \mat{Q}$ \updated{chosen via iterative tuning to achieve stable waypoint tracking with balanced control effort.}

\subsection{Observed Performance}

\Cref{fig:pose_and_command_comparison} illustrates the tracking and control behavior observed in both simulation and hardware. The followers successfully tracked the leader's communicated trajectory, maintaining formation within \qty{0.17}{\metre} in simulation and \qty{0.35}{\metre} on hardware. On the ATMOS platforms, a steady-state offset was visible across all agents, attributed to unmodeled disturbances such as floor unevenness and actuation asymmetries. These results highlight the ability of the BSK-ROS2 bridge to support seamless integration of modular autonomy stacks across platforms. Without changes to the controller implementation or architecture, the same \ac{ROS2}-based control nodes operate reliably in both high-fidelity simulation and hardware, validating the bridge's utility for spacecraft formation control and \ac{HITL} development workflows.

\updated{
\subsection{Communication Performance}

Stress tests were conducted with an AMD~Ryzen~5~5600X running Ubuntu~24.04 to assess the bridge's scalability and communication limits under varying simulation speeds, spacecraft counts, and topic rates. As shown in \Cref{tab:bridge_stress_test}, the bridge remained stable at nominal control frequencies, with saturation only under extreme loads.

\begin{table}[b]
\caption{Bridge communication performance}
\label{tab:bridge_stress_test}
\centering
\begin{tabular}{lcccc}
\toprule
\textbf{Sim speed} & \textbf{S/c} & \textbf{Target [Hz]} & \textbf{Achieved [Hz]} & \textbf{Std [ms]} \\
\midrule
1x    & 1   & 100     & 100   & 1.3 \\
1x    & 1   & 10\,000 & 4\,390  & 1.0 \\
100x  & 1   & 10\,000     & 4\,270    & 1.0 \\
1x    & 100 & 100     & 100   & 6.5 \\
100x  & 100 & 100       & 100    & 6.6 \\
100x  & 100 & 1\,000      & 270    & 2.4 \\
\bottomrule
\end{tabular}
\end{table}

It sustained 100\,Hz per agent with low jitter and reached about 4.4\,kHz in single-agent high-load tests. Similar limits under 100x accelerated simulation indicate CPU-bound throughput rather than timing constraints. With 100 spacecraft, it showed saturation near 270\,Hz per agent. These results confirm that the bridge comfortably supports typical spacecraft control frequencies.
}
% By launching multiple bridge instances with a corresponding Basilisk handler pair in the simulation, the load performance can be improved.

%% file: sections/Discussion_Conclusions.tex
\section{Discussion}

The bridge developed in this work has proven to be a practical enabler for the rapid deployment of modular autonomy stacks in both simulation and hardware environments. By decoupling the control logic from the simulation engine, the same \ac{ROS2} nodes used in Basilisk could be deployed on the ATMOS testbed with minimal changes. Specifically, the transition required only substituting Basilisk topics with those used by PX4, demonstrating the flexibility and portability of the proposed architecture. Such topic remappings can be easily automated. A key benefit of this approach is the ability to interact with the simulator in real-time without modifying its internal setup, which significantly lowers the barrier for incorporating advanced autonomy into existing Basilisk scenarios. The current setup was executed in an open-loop configuration, with no \ac{SITL} or \ac{HITL} feedback between Basilisk and the hardware. However, the bridge architecture inherently supports closed-loop setups, and enabling \ac{SITL} or \ac{HITL} is a promising avenue for future work.

The formation flying scenario presented here, while intentionally constrained to a planar 2D setting with one leader and two followers, demonstrates the core functionality and viability of the bridge. More advanced use cases can now be explored by leveraging Basilisk's extensive module library, which includes high-fidelity spacecraft dynamics, \ac{FSW} architectures, realistic sensors, actuator models, and environmental disturbances. This includes scenarios such as spacecraft docking, articulated manipulators, deployable structures, and flexible appendages. Combined with the \ac{ROS2} bridge, such capabilities can be coupled with real-time autonomy stacks, making them accessible for closed-loop testing, planning, and control.

\section{Conclusions}

This paper presents a lightweight and extensible communication bridge between the Basilisk Astrodynamics Simulation Framework and \ac{ROS2}. By enabling bidirectional data exchange without requiring modifications to Basilisk's internal architecture, the bridge allows real-time interaction between high-fidelity orbital dynamics and modular control stacks. Its effectiveness has been demonstrated in a leader–follower formation flying scenario using decentralized \ac{NMPC}, where the same \ac{ROS2}-based control system was deployed without modification in both Basilisk and the ATMOS 2D microgravity platform. The results highlight the bridge's utility in enabling seamless simulation-to-hardware transitions and supporting reproducible autonomy development workflows. Future work includes integration into \ac{SITL} and \ac{HITL} pipelines, \updated{scaling to larger constellations, and applications involving contact dynamics.} The open-source release of the bridge aims to accelerate adoption within the space robotics community and serve as a foundation for modular, scalable autonomy architectures in orbital applications.

%% file: References/channn_ref.bib
@article{scFormationControllerStateOfTheArt,
   author = {Raymond Kristiansen and Per Johan Nicklasson},
   doi = {10.1016/J.ACTAASTRO.2009.04.014},
   issn = {0094-5765},
   issue = {11-12},
   journal = {Acta Astronautica},
   pages = {1537-1552},
   publisher = {Pergamon},
   title = {Spacecraft formation flying: A review and new results on state feedback control},
   volume = {65},
   year = {2009},
}

@article{PRISMA-mission,
   author = {Per Bodin and Ron Noteborn and Robin Larsson and Thomas Karlsson and Simone D'Amico and Jean Sebastien Ardaens and Michel Delpech and Jean Claude Berges},
   issn = {00653438},
   journal = {Adv. Astronaut. Sci.},
   title = {The prisma formation flying demonstrator: Overview and conclusions from the nominal mission},
   volume = {144},
   year = {2012},
}

@article{PROBA-3-Mission,
   author = {Thomas V. Peters and João Branco and Diego Escorial and Lorenzo Tarabini Castellani and Alex Cropp},
   doi = {10.1016/j.actaastro.2014.01.010},
   journal = {Acta Astronautica},
   title = {Mission Analysis for {PROBA-3} Nominal Operations},
   volume = {102},
   year = {2014},
}

@article{basiliskIntro,
   author = {Patrick W. Kenneally and Scott Piggott and Hanspeter Schaub},
   doi = {10.2514/1.I010762},
   issn = {23273097},
   issue = {9},
   journal = {Journal of Aerospace Information Systems},
   title = {Basilisk: {A} flexible, scalable and modular astrodynamics simulation framework},
   volume = {17},
   year = {2020},
}

@inproceedings{Basilisk-ROS2-Bridge-1,
   author = {Kai Matsuka and Leo Zhang and Isabelle Ragheb and Christine Ohenzuwa and Soon-Jo Chung},
   booktitle = {AAS/AIAA Space Flight Mech.},
   title     = {High-fidelity Closed-Loop Simulation of Spacecraft Vision-Based Relative Navigation in {ROS2}},
   year = {2023},
}

@inproceedings{todorov2012mujoco,
  title={{MuJoCo}: {A} Physics Engine for Model-Based Control},
  author={Todorov, Emanuel and Erez, Tom and Tassa, Yuval},
  booktitle={Proc. IEEE/RSJ Int. Conf. Intell. Robots Syst. (IROS)},
  pages={5026--5033},
  year={2012},
  organization={IEEE},
  doi={10.1109/IROS.2012.6386109}
}

@conference{Bonilla:2025kh,
address = {Boston, {MA}},
author = {Juan {Garcia Bonilla} and Hanspeter Schaub},
booktitle = {AAS/AIAA Astrodyn. Specialist Conf.},
title = {{A} Message-Passing Simulation Framework for Generally Articulated Spacecraft Dynamics},
year = {2025}
}

@article{ROS2,
    author = {Steven Macenski and Tully Foote and Brian Gerkey and Chris Lalancette and William Woodall},
    title = {Robot operating system 2: {Design}, architecture, and uses in the wild},
    journal = {Science Robotics},
    volume = {7},
    number = {66},
    year = {2022},
    doi = {10.1126/scirobotics.abm6074},
}

@inproceedings{spaceROS,
   author = {Austin B. Probe and S. Will Chambers and Amalaye Oyake and Matthew Deans and Guillaume Brat and Nick Cramer and Brian Kempa and Brian Roberts and Kimberly Hambuchen},
   doi = {10.2514/6.2023-2709},
   booktitle = {Proc. AIAA SciTech Forum},
   title = {{Space} {ROS}: {An} Open-Source Framework for Space Robotics and Flight Software},
   year = {2023}
}

@misc{zmq,
   title = {ZeroMQ | Socket API},
   url = {https://zeromq.org/socket-api/}
}

@article{CW-Eqn,
   author = {William H. Clohessy and Robert S. Wiltshire},
   doi = {10.2514/8.8704},
   issue = {9},
   journal = {Journal of the Aerospace Sciences},
   title = {Terminal Guidance System for Satellite Rendezvous},
   volume = {27},
   year = {1960},
}

@mastersthesis{thomas-SFF-Thesis,
   author = {Ngai Nam Chan},
   school = {KTH Royal Institute of Technology},
   title = {Implementation of Controller Schemes for Multi-Agent Spacecraft Formation Flights via the Basilisk Simulation Framework},
   year = {2025}
}


%% file: References/eliaskra_ref.bib
@inproceedings{bualat_astrobee_2015,
	title = {Astrobee: {Developing} a free-flying robot for the {International} {Space} {Station}},
	shorttitle = {Astrobee},
	urldate = {2025-08-02},
	booktitle = {{AIAA} {SPACE} {Conf}.},
	author = {Bualat, Maria and Barlow, Jonathan and Fong, Terrence and Provencher, Chris and Smith, Trey},
	year = {2015},
}

@book{rawlings_model_2020,
	edition = {2nd Edition},
	title = {Model {Predictive} {Control}: {Theory}, {Computation}, and {Design}},
	urldate = {2025-08-04},
	publisher = {Nob Hill Publishing},
	author = {Rawlings, James B. and Mayne, David Q. and Diehl, Moritz M.},
	year = {2020},
}

@misc{roque_towards_2025,
	title = {Towards open-source and modular space systems with {ATMOS}},
	doi = {10.48550/arXiv.2501.16973},
	urldate = {2025-08-02},
	publisher = {arXiv},
	author = {Roque, Pedro and Phodapol, Sujet and Krantz, Elias and Lim, Jaeyoung and Verhagen, Joris and Jiang, Frank J. and Dörner, David and Mao, Huina and Tibert, Gunnar and Siegwart, Roland and Stenius, Ivan and Tumova, Jana and Fuglesang, Christer and Dimarogonas, Dimos V.},
	year = {2025},
	note = {arXiv:2501.16973 [cs]},
}

@techreport{fluckiger_astrobee_2018,
	title = {Astrobee robot software: {A} modern software system for space},
	shorttitle = {Astrobee {Robot} {Software}},
	urldate = {2025-08-02},
	institution = {NASA Ames Research Center},
	author = {Flückiger, Lorenzo and Browne, Kathryn and Coltin, Brian and Fusco, Jesse and Morse, Theodore and Symington, Andrew},
	year = {2018},
}

@inproceedings{kolvenbach_recent_2016,
	title = {Recent developments on {ORBIT}, a 3-{DoF} free-floating contact dynamics testbed},
	language = {en},
	booktitle = {in {Proc}. {Int}. {Symp}. {Artif}. {Intell}., {Robot}. {Autom}. {Space} (i-{SAIRAS})},
	author = {Kolvenbach, H and Wormnes, K},
	year = {2016},
}

@inproceedings{meier_px4_2015,
	title = {{PX4}: {A} node-based multithreaded open source robotics framework for deeply embedded platforms},
	shorttitle = {{PX4}},
	doi = {10.1109/ICRA.2015.7140074},
	urldate = {2025-08-02},
	booktitle = {Proc. {IEEE} {Int}. {Conf}. {Robot}. {Autom}. ({ICRA})},
	author = {Meier, Lorenz and Honegger, Dominik and Pollefeys, Marc},
	year = {2015},
	pages = {6235--6240},
}

@inproceedings{barrett_demonstration_2009,
	title = {Demonstration of technologies for autonomous micro-satellite assembly},
	isbn = {978-1-60086-980-8},
	doi = {10.2514/6.2009-6504},
	language = {en},
	urldate = {2025-08-03},
	booktitle = {{AIAA} {SPACE} {Conf}.},
	author = {Barrett, Tim and Schultz, Steve and Bezouska, William and Aherne, Michael},
	year = {2009},
}

@article{verschueren_acadosmodular_2022,
	title = {Acados—{A} modular open-source framework for fast embedded optimal control},
	volume = {14},
	issn = {1867-2957},
	doi = {10.1007/s12532-021-00208-8},
	language = {en},
	number = {1},
	urldate = {2025-08-05},
	journal = {Math. Program. Comput.},
	author = {Verschueren, Robin and Frison, Gianluca and Kouzoupis, Dimitris and Frey, Jonathan and Duijkeren, Niels van and Zanelli, Andrea and Novoselnik, Branimir and Albin, Thivaharan and Quirynen, Rien and Diehl, Moritz},
	year = {2022},
	pages = {147--183},
}

@inproceedings{kramlich_validation_2023,
	address = {Big Sky, Montana},
	title = {Validation of the guidance, navigation, and control ({GNC}) architecture for {RPO} missions using {Basilisk}-{CFS} architecture},
	language = {en},
	urldate = {2025-08-05},
	booktitle = {{AAS}/{AIAA} {Astrodyn}. {Specialist} {Conf}.},
	author = {Kramlich, Ryan and Huun, Jack and Garcia, Axel and Cohen, Arielle and Gaston, Tyler},
	year = {2023},
}

@inproceedings{koenig_design_2004,
	title = {Design and use paradigms for {Gazebo}, an open-source multi-robot simulator},
	volume = {3},
	doi = {10.1109/IROS.2004.1389727},
	urldate = {2025-08-03},
	booktitle = {Proc. {IEEE}/{RSJ} {Int}. {Conf}. {Intell}. {Robots} {Syst}. ({IROS})},
	author = {Koenig, N. and Howard, A.},
	year = {2004},
	pages = {2149--2154},
}

@misc{imatix_zeromq_nodate,
	title = {{ZeroMQ}},
	url = {https://zeromq.org/},
	urldate = {2025-08-03},
	author = {{iMatix}},
}

@misc{avs_laboratory_basilisk_nodate,
	title = {Basilisk {Astrodynamics} {Simulation} {Framework}},
	url = {https://avslab.github.io/basilisk/},
	urldate = {2025-08-03},
	author = {{AVS Laboratory}},
}

@misc{nvidia_corporation_isaac_nodate,
	title = {Isaac {Sim}},
	url = {https://developer.nvidia.com/isaac-sim},
	urldate = {2025-08-03},
	journal = {NVIDIA Developer},
	author = {{NVIDIA Corporation}},
}


%% file: References/ieeectl.bib
@IEEEtranBSTCTL{IEEEexample:BSTcontrol,
  CTLuse_forced_etal       = "yes",
  CTLmax_names_forced_etal = "6",
  CTLnames_show_etal       = "1"
}
